\documentclass[final]{hld2023}

\usepackage[utf8]{inputenc} % allow utf-8 input
\usepackage[T1]{fontenc}    % use 8-bit T1 fonts
\usepackage{hyperref}       % hyperlinks
\usepackage{booktabs}       % professional-quality tables
\usepackage{amsfonts}       % blackboard math symbols
\usepackage{nicefrac}       % compact symbols for 1/2, etc.
\usepackage{microtype}      % microtypography
\usepackage{xcolor}         % colors
\usepackage{caption}
\usepackage{subcaption}
\usepackage{bm} % for bold
\usepackage{physics} % for grad

\title{The instabilities of large learning rate training: a loss landscape view}

\hldauthor{%
\Name{Lawrence Wang} \Email{lawrence.wang@exeter.ox.ac.uk}\\
\addr {Dept. of Engineering, University of Oxford} 
\AND
\Name{Stephen J. Roberts} \Email{sjrob@robots.ox.ac.uk}\\
\addr {Dept. of Engineering, University of Oxford} }

\begin{document}

\newcommand{\neff}{$N_\mathrm{eff}$}

\newcommand{\lammax}{$\lambda_\mathrm{max}$}
\newcommand{\lamneg}{$\lambda^{-}_\mathrm{max}$}
\newcommand{\vmax}{$v_\mathrm{max}$}
\newcommand{\grassd}{$\hat{d_m}$}

\newcommand{\lr}{$\eta$}
\newcommand{\rotins}{$\overline{S(v)}$}
\newcommand{\gins}{$\overline{S(g)}$}
\newcommand{\hessvar}{$\sigma_{\lambda_\mathrm{bulk}}$}
\newcommand{\EF}{\textit{effective dimensionality}}
\newcommand{\dof}{\textit{d.o.f.}}
\newcommand{\dofs}{\textit{d.o.f.}s}

\newcommand{\sane}{SANE}
\newcommand{\eos}{\textit{Edge of Stability}}
\newcommand{\gf}{\textit{smooth}}
\newcommand{\Gf}{\textit{Gradient flow}}
\newcommand{\alf}{$\alpha$}
\newcommand{\fcite}{\textcolor{red}{find cite}}
\newcommand{\redhess}{$\mathcal{H}^\mathrm{r}_k$}
\newcommand{\todo}{\textcolor{red}{Needs work.}}

\newcommand{\app}{\textbf{Appendix}}

\maketitle
\begin{abstract}

Modern neural networks are undeniably successful. Numerous works study how the curvature of loss landscapes can affect the quality of solutions. In this work we study the loss landscape by considering the Hessian matrix during network training with large learning rates - an attractive regime that is (in)famously unstable. We characterise the instabilities of gradient descent, and we observe the striking phenomena of \textit{landscape flattening} and \textit{landscape shift}, both of which are intimately connected to the instabilities of training.

\end{abstract}

\section{Introduction}\label{section:intro}

Deep neural networks are undeniably successful across many tasks \cite{brown2020language, vaswani2017attention}. It is widely accepted that variations in the curvature of weight-space will affect model performance, where solutions in flatter regions generalise better to unseen data \cite{Hochreiter1997, keskar2017, hoffer2018train}. Naturally, the study of second-order characteristics of the \textit{loss landscape} through the Hessian (of the loss w.r.t. neural weights) uncovers geometrical details to quantify curvature and can allow efficient optimisation up to a local quadratic approximation \cite{Ghorbani2019, li2018visualizing}. Estimating the curvature of the loss basin through the Hessian has led to recent methods that actively seek flat regions in weight space during optimisation \cite{Foret2020, izmailov2019swa}.  

Recent studies call into question the link between landscape curvature and generalisation by showing manipulations of curvature through the learning rate \cite{Kaur2022} or via regularisation \cite{granziol2020flatness} without the purported changes to generalisation as a result of sharpness. Additionally, \citet{cohen2022gradient} showed the performance of models can continue to improve with unstable learning rates, despite instability in training, which challenges existing wisdom \cite{lecun1992} (i.e. to avoid divergence) for learning rate selection. This encourages a large learning rate regime for training, which trades the non-monotonicity of loss for more effective steps across weight space.

In this work, we study gradient descent with large learning rates through the orientation of the loss landscape. We employ a bulk-outlier decomposition of the Hessian, where each outlier (sharp) eigendirection of the Hessian represents an "effective" parameter of the model. "Effective" parameters control important degrees of freedom (d.o.f.) of the model solution. We use these insights to study the instabilities of large learning rate gradient descent. \textbf{Our contributions are as follows:} 

\begin{enumerate}
    \item We characterise the phases of gradient descent instabilities 
    \item \textit{Landscape flattening:} The worst-case (maximum) sharpness of solutions decreases as the number of instabilities increases
    \item \textit{Landscape shift:} The loss landscape shifts at each training instability. The orientation of the landscape, which governs the performance of the solutions, can remain similar at low $\eta$s while high $\eta$s encourage the exploration of new solutions
    \item To promote openness and reproducibility of research, we make our code available\footnote{https://github.com/lawrencewang94/gd-instabs}
\end{enumerate}

We introduce the notation and the motivation for our work in Section \ref{section:background}, and we present our main findings in Section \ref{section:main}. We offer a brief discussion and conclude in Section \ref{section:discussion}.

\section{Methods \& Motivation}\label{section:background}

\noindent \textbf{Notation.} In this work we consider supervised classification where $\bm{x_i} = \{x_i, y_i\}$ constitutes an input-label pair. The predictions are modeled with a deep neural net with weights $\theta \in \mathbb{R}^{N} = \Theta$ to obtain a prediction function $\hat{y}_i = f_\theta(x_i)$. The loss function over the data set is $\mathcal{L}_{\theta} = \mathbb{E}_{\bm{x_i}} L_{\theta}(\bm{x_i})$, where $L_{\theta}$ is the (cross-entropy) loss between the prediction $\hat{y}_i$ and the true label $y_i$. So, we can write the gradient $g = \grad_{\theta} \mathcal{L}_{\theta}$ and the Hessian $\mathcal{H}=\grad^2_{\theta} \mathcal{L}_{\theta}$. We order the eigenvalues of the Hessian in descending order: $\lambda_\mathrm{max}=\lambda_1>\lambda_2>...>\lambda_N=\lambda^{-}_\mathrm{max}$, and $v_{n}$ is the $n$-th eigenvector. \\

\noindent \textbf{Divergence of gradient descent.} Suitable learning rates for gradient descent naturally differ depending on the geometry of weight space, which is influenced by factors such as the dataset and the choice of architecture. Optimisers, such as ADAM \cite{kingma2017adam}, incorporate automatic preconditioning, but finding the right learning rates remains an empirical endeavour in practice, where too-high learning rates can lead to model divergence. For a convex quadratic function $f(x) = \frac{1}{2}\textbf{x}^T\textbf{A}\textbf{x}+\textbf{b}^Tx+c$, gradient descent with learning rate $\eta$ will diverge \textit{iff} any eigenvalue of $\textbf{A}$ exceeds the threshold $2/\eta$ \cite{Nesterov2013}. This bound is also known as the \eos \cite{cohen2022gradient}, and it is conventionally used as an upper bound for $\eta$ to prevent divergence of loss \cite{lecun1992, granziol2020lr}. However, \citet{cohen2022gradient} have shown that despite instabilities, gradient descent can continue to decrease the objective function consistently over long timescales. Alternatively, \citet{lewkowycz2020large} finds a \textit{catapult} regime of learning rates where gradient descent is unstable but does not diverge. In this regime, the model eventually gets "catapulted" into a region with low sharpness. These observations support using large learning rates that are \textit{unstable} to take larger steps across weight space and find better solutions, challenging existing stability theory which recommends $\eta<2/\lambda_\mathrm{max}$ to guarantee non-divergence of loss. We differentiate the learning rate regimes that fall under or exceed the \textit{stability} limit as the \gf\ and the \textit{unstable} regimes respectively. \\

\noindent \textbf{Outlier-bulk Hessian decomposition.} Recent studies on the structure of the Hessian \cite{granziol2020lr, papyan2019measurements, papyan2019spectrum, papyan2020traces, sagun2017eigenvalues} report a consistent separation of the \textit{outliers} from the \textit{bulk} of the spectrum. Through deflation techniques \cite{papyan2019spectrum}, \citet{papyan2020traces} presented empirical evidence that the spectral \textit{outliers} can be attributed to $\mathcal{G}$ and the \textit{bulk} to $\mathcal{E}$, from the genralised Gauss-Newton decomposition $\mathcal{H}=\mathcal{G}+\mathcal{E}$, where $\mathcal{G}$ is the \textit{Fisher Information Matrix}. 
In section \ref{section:main}, we will refer to a generic \textit{outlier}-\textit{bulk} decomposition of the Hessian inspired by these results:
\begin{align}
    \mathcal{H} &= \mathcal{H}_\mathrm{out} + \mathcal{H}_\mathrm{bulk} = V_{o}W_{o}V_{o}^{T}+\mathcal{H}_\mathrm{bulk}  \label{eqn:hess_decomp}
\end{align}
where $\mathcal{H}_\mathrm{out}$, $\mathcal{H}_\mathrm{bulk}$ are outlier/bulk components and $V_{o}W_{o}V_{o}^T$ the eigen-decomposition of $H_\mathrm{out}$. $\mathcal{H}_\mathrm{out}$ is assumed to positive definite, in typical empirical Hessians 
a thin band of small negative eigenvalues exists for empirical $\mathcal{H}$ which we associate with $\mathcal{H}_\mathrm{bulk}$. Note that $\lambda_n=\lambda^{-}_\mathrm{max}<0$, and $|\lambda^{-}_\mathrm{max}|$ is used to estimate the variance of the bulk Hessian. \\

\noindent \textbf{Loss landscapes.} We study the Hessian to uncover the geometrical properties of the loss landscape (loss-weight space) in which optimisation is conducted. Flat (wide) regions in the loss landscape are distinguished from sharp (narrow) regions if the objective function changes slowly to shifts in the parameters. It is widely believed that flat solutions generalise better to unseen data. \citet{Hochreiter1997} provided justification for this connection through the minimum description length framework, suggesting that flat minima permit the greatest compression of data. \citet{djcmThesis} showed, from a Bayesian perspective, that flat minima can be the consequence of an Occam's razor penalty. To estimate curvature, we adopt the standard approach in the literature which uses a quadratic approximation in the local weight space and equate the top eigenvalue (\lammax) of the Hessian to the sharpness (inverse-flatness) of the loss landscape curvature. 

The orientation of loss, through "effective parameters", is approximated by the top-$m$ sharpest eigenvectors, which represent the directions most informed by the data. In the \app, we show each "effective" parameter controls a significant degree of freedom, each of which corresponds to good model performance. The number of "effective" parameters is used in a Bayesian setting as a proxy to the effective dimensionality of models, while \citet{maddox2020rethinking} and \citet{sane} demonstrate its effectiveness as a tool for model comparison.  \\

\noindent \textbf{Similarity of landscapes.} We can approximate the similarity of \textit{loss basins} via the informative (outlier) eigenvectors, since the remaining eigen-directions are likely degenerate \cite{maddox2020rethinking}. Let $V^{*} \in \mathcal{R}^{m \times N}$ be the matrix of informative eigenvectors, we can compare the similarity of $m$-dimensional linear subspaces defined by $V^{*}$ with \textit{Grassmanian distance}, $d_m$ \cite{ye2016schubert}. Since the $V^{*}$s are orthonormal bases, we utilise singular value decomposition (SVD) to get:
\begin{align*}
    V^{*}_p (V^{*}_q)^T &= \mathcal{U}\Sigma \mathcal{V}^{T} &    \cos (\phi_i) &= \sigma_i \\
    d_m(V^{*}_p, V^{*}_q) &= \left( \sum ^{m} _{i=1} \phi_i^2 \right) ^{\frac{1}{2}} &    \hat{d_m}(V^{*}_p, V^{*}_q) &= \frac{d_m(V^{*}_p, V^{*}_q)}{\sqrt{m}\pi/2}
\end{align*}
where $\sigma_i$ are the eigenvalues of $\Sigma$. We scale $d_m$ by its maximum value (determined by $m$) and use a cosine to get $\hat{d_m} \in [0, 1]$. For $m=1$, $\cos (\hat{d_m} \frac{\pi}{2})$ is equal to the cosine similarity function, and $d_m$ can be extended to the case where $m_p \neq m_q$ \cite{ye2016schubert}. In the following sections, we compute the misalignment of vector spaces $S(p,q) = 1 - \cos ( \frac{\pi}{2} \hat{d_m}(p, q)).$ 
\\

\noindent \textbf{Computation.} The computation and storage of the full Hessian is expensive. We take advantage of existing auto-differentiation libraries \cite{jax2018github} to obtain the Hessian Vector Product (HVP) $\mathcal{H}v = \grad_\theta((\grad_\theta \mathcal{L})v) $, using Pearlmutter's trick \cite{pearlmutter}. We then use the HVP to compute the eigenvector-value pairs, $(\lambda_i,v_i)$, with the Krylov-based Lanczos iteration method \cite{Lanczos:1950zz}. See \app\ for details. 

\begin{figure}[t]
\centering
\includegraphics[width=1.0\textwidth]{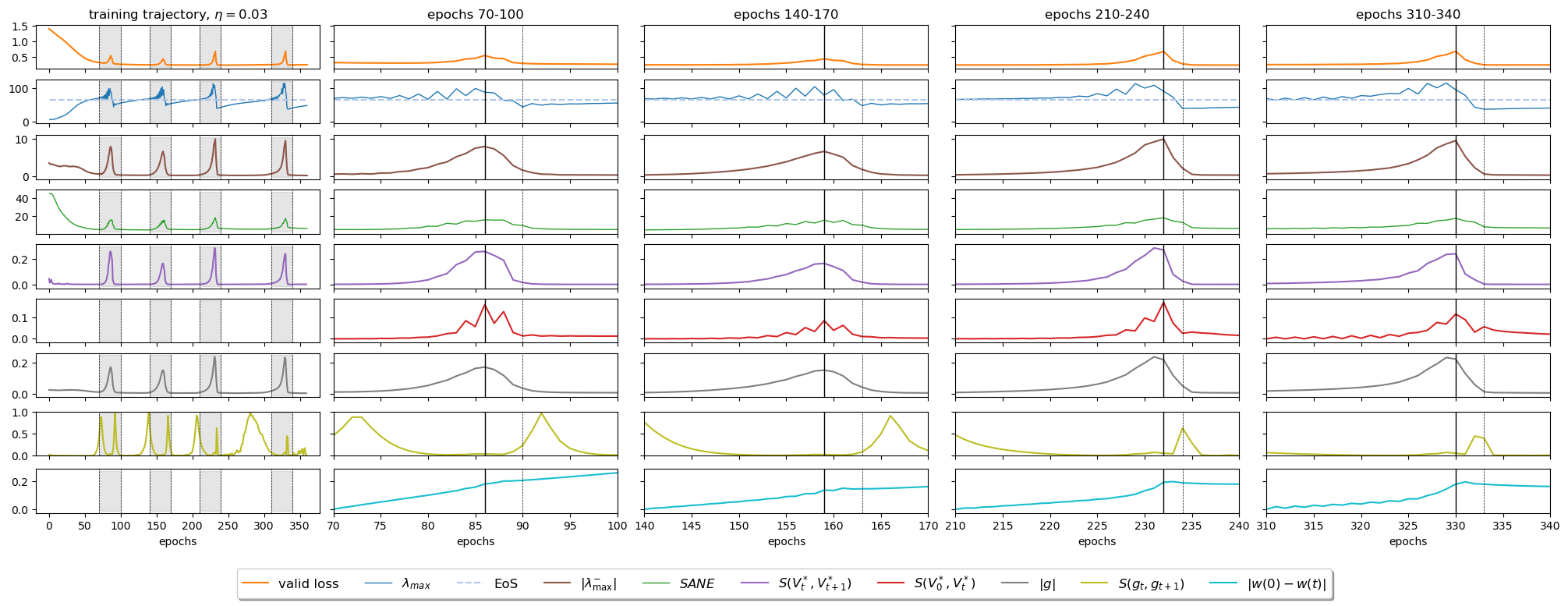}
\caption{\textbf{Instabilities of large learning-rate gradient descent.} Rows are ordered from top to bottom in the \textbf{legend}. The leftmost plot shows the whole trajectory and unstable epochs are highlighted in subplots.}
\label{fig:phases}
\end{figure}

\section{Main Experiments}\label{section:main}
Our experiments are conducted on fashionMNIST \cite{xiao2017fashionmnist}, a small benchmark for image classification. \\

\begin{figure}[b]
\centering
\includegraphics[width=0.9\textwidth]{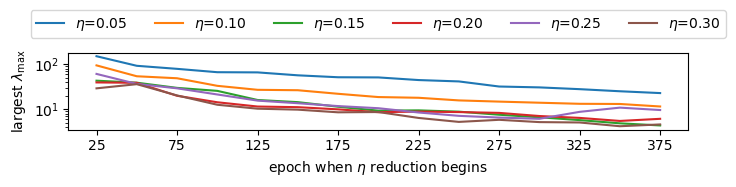}
\caption{\textbf{Delaying $\eta$ reduction reduces the worst-case \lammax\ of solutions.} }
\label{fig:breakaway}
\end{figure}

\noindent \textbf{Phases of instability.} In Section \ref{section:background}, we introduced an outlier-bulk ($W_o$, $V_o$, $H_\mathrm{bulk}$) decomposition of the Hessian (Eq. \ref{eqn:hess_decomp}). The interplay of the Hessian components (\textbf{sharpness} $W_o$, \textbf{orientation} $V_o$, and \textbf{bulk} $H_\mathrm{bulk}$) uncovers insights into the \textit{instabilities} of gradient descent. Fig. \ref{fig:phases} plots an \textit{unstable} training trajectory, where we visualise movements in the loss among measures of the Hessian and gradients. The instabilities manifest as spikes in the trajectory but are non-divergent since the model metrics return to pre-spike values. Distinct \textbf{heating} and \textbf{cooling} phases are observed, and there exists a sharp demarcation from heating to cooling that is marked by the synchronous changes in loss, landscape similarity, gradient norm, and the variance of the bulk hessian. Note that \lammax\ appears to be roughly (as opposed to sharply) aligned with this phase change, which makes it a much less accurate signal than other measures we consider. In contrast, the boundary from cooling to heating is less clear. The length and behaviour of these phases are not symmetrical. \\

\noindent \textbf{Landscape flattening.} What happens to the maximum sharpness of solutions if extended along a stable trajectory? In Fig. \ref{fig:breakaway}, we plot the worst-case \lammax\ of solutions when we reduce $\eta$ at different points along the training trajectory to remove the stability constraint. As we vary the timing at which $\eta$ reduction begins, we observe that on large timescales the worst-case \lammax\ of solutions decreases as $\eta$ reduction is delayed. While most intervals contained an instability, the final few data points for $\eta=0.25$ and $\eta=0.30$, exhibited an increase in worst-case \lammax\ which could describe the effects of $\eta$ reduction in stable $\eta$ regimes. Since we reduce $\eta$ to equal the benchmark \gf\ trajectory, this relationship isn't present in the \gf\ regime. The evidence suggests that \textbf{the maximum sharpness of solutions decreases as the number of instabilities increases}. \\

\begin{figure}[t]
\centering
 \begin{subfigure}[b]{0.24\textwidth}
     \centering
     \includegraphics[height=3.3cm]{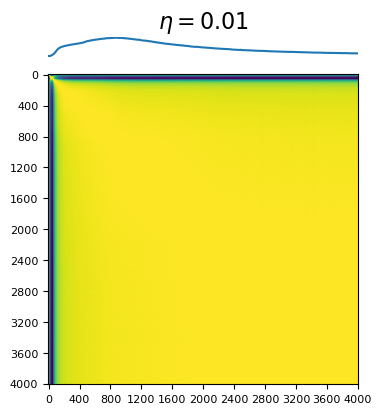}
     \caption{$\eta=0.01$}
     \label{fig:sim1}
 \end{subfigure}
 % \hfill
 \begin{subfigure}[b]{0.24\textwidth}
     \centering
     \includegraphics[height=3.3cm]{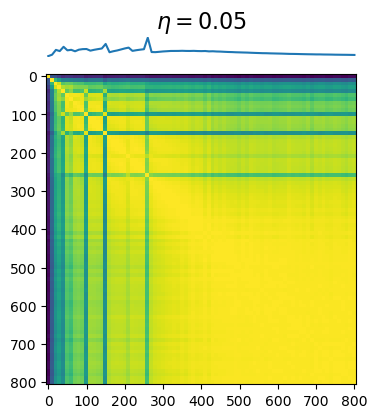}
     \caption{$\eta=0.05$}
     \label{fig:sim2}
 \end{subfigure}
 % \hfill
 \begin{subfigure}[b]{0.24\textwidth}
     \centering
     \includegraphics[height=3.3cm]{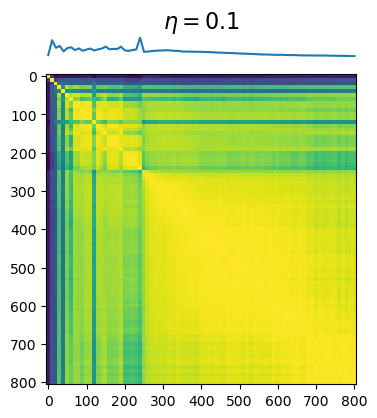}
     \caption{$\eta=0.10$}
     \label{fig:sim3}
 \end{subfigure}
% \centering
\begin{subfigure}[b]{0.24\textwidth}
    \centering
    \includegraphics[height=3.3cm]{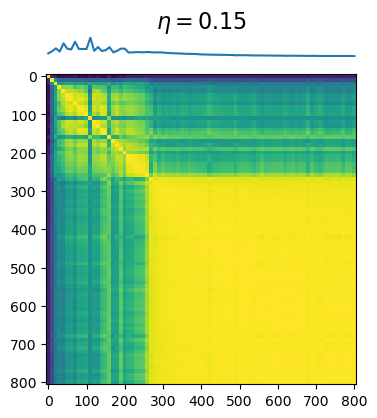}
    \caption{$\eta=0.15$}
    \label{fig:sim4}
\end{subfigure}

\begin{subfigure}[b]{0.24\textwidth}
    \centering
    \includegraphics[height=3.3cm]{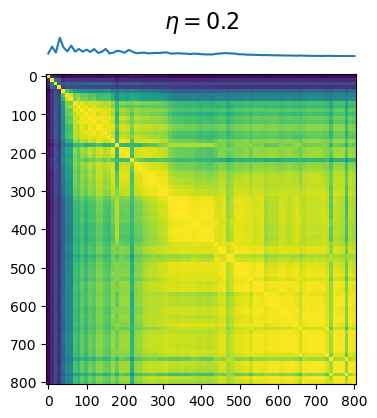}
    \caption{$\eta=0.20$}
    \label{fig:sim5}
\end{subfigure}
\begin{subfigure}[b]{0.24\textwidth}
    \centering
    \includegraphics[height=3.3cm]{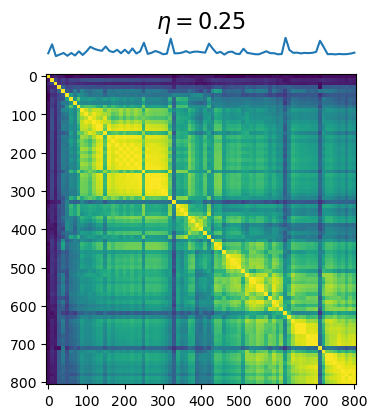}
    \caption{$\eta=0.25$}
    \label{fig:sim6}
\end{subfigure}
\begin{subfigure}[b]{0.24\textwidth}
    \centering
    \includegraphics[height=3.3cm]{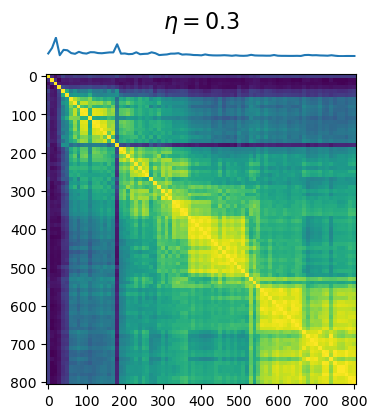}
    \caption{$\eta=0.30$}
    \label{fig:sim7}
\end{subfigure}
\caption{\textbf{Landscape similarity of trajectories at different learning rates}. Plots of $S(V^*_p, V^*_q)$ as a (symmetrical) heatmap using \textit{colourmap viridis} (yellow is more similar) \cite{Hunter:2007}. \lammax\ trajectories are plotted above. As $\eta$ increases, the regions of landscape similarity become more local. }
\label{fig:sim}
\end{figure}

\noindent \textbf{Landscape shift.} We measure the similarity of \textit{loss landscapes} as they change between instabilities. In Fig. \ref{fig:phases}, the increase in $w(t_0)-w(t)$ suggests that the loss basins are well-separated before and after instabilities. However, the closeness of $S(V^*_0, V^*_{t}$ shows that these basins are dominated by the same "effective" parameters. We show in the \app\ that, in many cases, using $m=1$, leads to poor similarities, which suggests that there exists $m \ll n$ where $|S(V^*_{t_0},V^*_t)|<\epsilon$ for some $\epsilon$. 

We turn to the effect of $\eta$ on landscape shift. In Fig. \ref{fig:sim}, we plot the $S(V^*_p, V^*_q)$ where $p,q$ are taken from the entire trajectory. The initialisation is identical for each trajectory, and for this particular initialisation, the \textit{unstable} regime of learning begins at $\eta=0.03$. Using the \gf\ $\eta=0.01$ trajectory as a baseline, we see that the trajectory of low $\eta$ are largely similar as we enter the \textit{unstable} regime. Eventually, the windows of similarity become more local when $\eta$ is large, showing that the "effective" parameters of the Hessians change across instabilities. Given the connection between "effective" parameters and solutions, this suggests that \textbf{large learning rates encourage the exploration of new solutions.} 

\begin{figure}[t]
\centering
\includegraphics[width=0.8\textwidth]{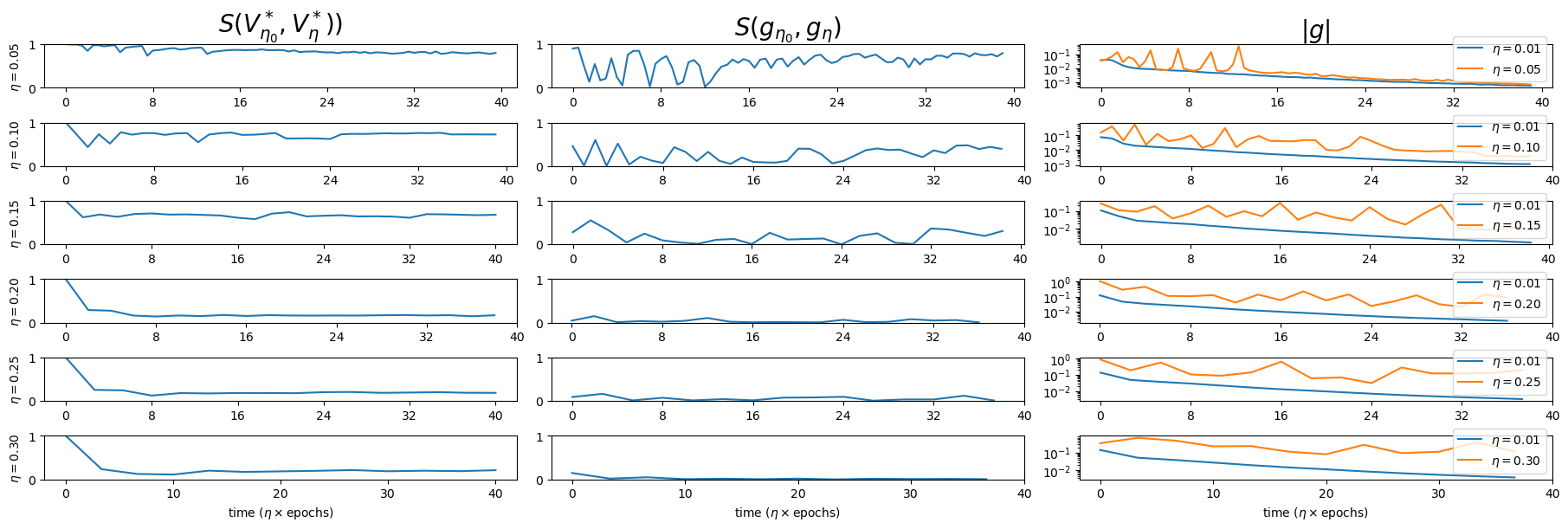}
\caption{\textbf{Comparison of training trajectories to the $\eta=0.01$ smooth baseline.} As $\eta$ is increased, the trajectories become less similar with the \gf\ baseline.}

\label{fig:gfsim}
\end{figure}

Alternatively, we use $\eta=0.01$ as a baseline to compare trajectories across learning rates via gradient and Hessian similarities. These results are plotted in Fig. \ref{fig:gfsim}. We note the extraordinary similarity between $\eta=0.05$ and $\eta=0.01$, where the Hessian rotation and gradient directions are largely similar, and $|g|$ is identical when it recovers from instabilities. As $\eta$ is increased, the trajectories become less similar to the baseline, until $\eta=0.20$ beyond which the trajectories become extremely dissimilar. The similarity of trajectories with $\eta$ to the \gf\ baseline supports the view that \textbf{low learning rates tend to preserve existing solutions}. 

\section{Discussion} \label{section:discussion}

In this work, we describe the phases of gradient descent instabilities. Our observations highlight \textit{landscape flattening} and \textit{landscape shift} across different learning rates. As we observe \textit{landscape flattening} and \textit{landscape shift}, the constancy of landscape orientation and validation loss provides further evidence for the connection between "effective" parameters and generalisation. Intuitively, we can view \textit{landscape shifts} at low $\eta$ to be a re-ordering of "effective" parameters, which is ineffectual to model performance as $V^*$ is not shifted. 

Our experiments build toward a body of studies characterising the behaviour of gradient descent at large learning rates as the relationship between large learning rates and generalisation attracts increasing attention. Following \citet{cohen2022gradient}, our work further shows that there exist large $\eta$s that do not cause divergence but may introduce significant changes to the orientation of the loss landscape where consistent a decrease in the training objective over long timescales may not be guaranteed. \textit{Progressive sharpening} \cite{cohen2022gradient} offers an explanation for the \textbf{heating} phase, the mechanisms of the cooling phase are less well studied and our experiments add to the body of evidence for its understanding. \citet{lewkowycz2020large} shows that for low but unstable $\eta$s, optimisers are "catapulted" around during training instabilities, which is supported by \textit{landscape flattening}. 

Our work presents an initial foray into the description of learning dynamics with large learning rates, but a complete understanding has some way to go. A better taxonomy and understanding of these regimes could speed up optimisation for practitioners using gradient methods for deep networks. We exploited the connection between Hessian eigen-directions to problem-specific degrees-of-freedom, so strengthening this connection would be beneficial toward generalisation. We see information compression and the variance of the bulk Hessian playing crucial, but under-explored, roles in this connection. 

\newpage

\bibliography{my_bibs}

\begin{thebibliography}{33}
\providecommand{\natexlab}[1]{#1}
\providecommand{\url}[1]{\texttt{#1}}
\expandafter\ifx\csname urlstyle\endcsname\relax
  \providecommand{\doi}[1]{doi: #1}\else
  \providecommand{\doi}{doi: \begingroup \urlstyle{rm}\Url}\fi

\bibitem[Bradbury et~al.(2018)Bradbury, Frostig, Hawkins, Johnson, Leary,
  Maclaurin, Necula, Paszke, Vander{P}las, Wanderman-{M}ilne, and
  Zhang]{jax2018github}
James Bradbury, Roy Frostig, Peter Hawkins, Matthew~James Johnson, Chris Leary,
  Dougal Maclaurin, George Necula, Adam Paszke, Jake Vander{P}las, Skye
  Wanderman-{M}ilne, and Qiao Zhang.
\newblock {JAX}: composable transformations of {P}ython+{N}um{P}y programs,
  2018.
\newblock URL \url{http://github.com/google/jax}.

\bibitem[Brown et~al.(2020)Brown, Mann, Ryder, Subbiah, Kaplan, Dhariwal,
  Neelakantan, Shyam, Sastry, Askell, Agarwal, Herbert-Voss, Krueger, Henighan,
  Child, Ramesh, Ziegler, Wu, Winter, Hesse, Chen, Sigler, Litwin, Gray, Chess,
  Clark, Berner, McCandlish, Radford, Sutskever, and Amodei]{brown2020language}
Tom~B. Brown, Benjamin Mann, Nick Ryder, Melanie Subbiah, Jared Kaplan,
  Prafulla Dhariwal, Arvind Neelakantan, Pranav Shyam, Girish Sastry, Amanda
  Askell, Sandhini Agarwal, Ariel Herbert-Voss, Gretchen Krueger, Tom Henighan,
  Rewon Child, Aditya Ramesh, Daniel~M. Ziegler, Jeffrey Wu, Clemens Winter,
  Christopher Hesse, Mark Chen, Eric Sigler, Mateusz Litwin, Scott Gray,
  Benjamin Chess, Jack Clark, Christopher Berner, Sam McCandlish, Alec Radford,
  Ilya Sutskever, and Dario Amodei.
\newblock Language models are few-shot learners, 2020.

\bibitem[Cohen et~al.(2022)Cohen, Kaur, Li, Kolter, and
  Talwalkar]{cohen2022gradient}
Jeremy~M. Cohen, Simran Kaur, Yuanzhi Li, J.~Zico Kolter, and Ameet Talwalkar.
\newblock Gradient descent on neural networks typically occurs at the edge of
  stability, 2022.

\bibitem[Foret et~al.(2020)Foret, Research, Research, and Blueshift]{Foret2020}
Pierre Foret, Ariel Kleiner~Google Research, Hossein Mobahi~Google Research,
  and Behnam~Neyshabur Blueshift.
\newblock Sharpness-aware minimization for efficiently improving
  generalization.
\newblock 10 2020.
\newblock \doi{10.48550/arxiv.2010.01412}.
\newblock URL \url{https://arxiv.org/abs/2010.01412v3}.

\bibitem[Fukushima(1975)]{relu}
Kunihiko Fukushima.
\newblock Cognitron: A self-organizing multilayered neural network.
\newblock \emph{Biol. Cybern.}, 20\penalty0 (3–4):\penalty0 121–136, sep
  1975.
\newblock ISSN 0340-1200.
\newblock \doi{10.1007/BF00342633}.
\newblock URL \url{https://doi.org/10.1007/BF00342633}.

\bibitem[Ghorbani et~al.(2019)Ghorbani, Krishnan, and Xiao]{Ghorbani2019}
Behrooz Ghorbani, Shankar Krishnan, and Ying Xiao.
\newblock An investigation into neural net optimization via {H}essian
  eigenvalue density.
\newblock \emph{36th International Conference on Machine Learning, ICML 2019},
  2019-June:\penalty0 4039--4052, 1 2019.
\newblock \doi{10.48550/arxiv.1901.10159}.
\newblock URL \url{https://arxiv.org/abs/1901.10159v1}.

\bibitem[Granziol(2020)]{granziol2020flatness}
Diego Granziol.
\newblock Flatness is a false friend, 2020.

\bibitem[Granziol et~al.(2020)Granziol, Zohren, Roberts, and
  Lacoste-Julien]{granziol2020lr}
Diego Granziol, Stefan Zohren, Stephen Roberts, and Simon Lacoste-Julien.
\newblock Learning rates as a function of batch size: A random matrix theory
  approach to neural network training.
\newblock \emph{Journal of Machine Learning Research}, 1:\penalty0 1--48, 2020.

\bibitem[Hochreiter and Schmidhuber(1997)]{Hochreiter1997}
Sepp Hochreiter and Jürgen Schmidhuber.
\newblock {Flat Minima}.
\newblock \emph{Neural Computation}, 9\penalty0 (1):\penalty0 1--42, 01 1997.
\newblock ISSN 0899-7667.
\newblock \doi{10.1162/neco.1997.9.1.1}.
\newblock URL \url{https://doi.org/10.1162/neco.1997.9.1.1}.

\bibitem[Hoffer et~al.(2018)Hoffer, Hubara, and Soudry]{hoffer2018train}
Elad Hoffer, Itay Hubara, and Daniel Soudry.
\newblock Train longer, generalize better: closing the generalization gap in
  large batch training of neural networks, 2018.

\bibitem[Hunter(2007)]{Hunter:2007}
J.~D. Hunter.
\newblock Matplotlib: A 2d graphics environment.
\newblock \emph{Computing in Science \& Engineering}, 9\penalty0 (3):\penalty0
  90--95, 2007.
\newblock \doi{10.1109/MCSE.2007.55}.

\bibitem[Izmailov et~al.(2019)Izmailov, Podoprikhin, Garipov, Vetrov, and
  Wilson]{izmailov2019swa}
Pavel Izmailov, Dmitrii Podoprikhin, Timur Garipov, Dmitry Vetrov, and
  Andrew~Gordon Wilson.
\newblock Averaging weights leads to wider optima and better generalization,
  2019.

\bibitem[Kaur et~al.(2022)Kaur, Cohen, and Lipton]{Kaur2022}
Simran Kaur, Jeremy Cohen, and Zachary~C Lipton.
\newblock On the maximum {H}essian eigenvalue and generalization.
\newblock 2022.

\bibitem[Keskar et~al.(2017)Keskar, Mudigere, Nocedal, Smelyanskiy, and
  Tang]{keskar2017}
Nitish~Shirish Keskar, Dheevatsa Mudigere, Jorge Nocedal, Mikhail Smelyanskiy,
  and Ping Tak~Peter Tang.
\newblock On large-batch training for deep learning: Generalization gap and
  sharp minima, 2017.

\bibitem[Kingma and Ba(2017)]{kingma2017adam}
Diederik~P. Kingma and Jimmy Ba.
\newblock Adam: A method for stochastic optimization, 2017.

\bibitem[Krizhevsky and Hinton(2009)]{krizhevsky2009learning}
Alex Krizhevsky and Geoffrey Hinton.
\newblock Learning multiple layers of features from tiny images.
\newblock Technical Report~0, University of Toronto, Toronto, Ontario, 2009.

\bibitem[Krizhevsky et~al.(2012)Krizhevsky, Sutskever, and Hinton]{alexnet}
Alex Krizhevsky, Ilya Sutskever, and Geoffrey~E. Hinton.
\newblock Imagenet classification with deep convolutional neural networks.
\newblock In \emph{Proceedings of the 25th International Conference on Neural
  Information Processing Systems - Volume 1}, NIPS'12, page 1097–1105, Red
  Hook, NY, USA, 2012. Curran Associates Inc.

\bibitem[Lanczos(1950)]{Lanczos:1950zz}
Cornelius Lanczos.
\newblock {An iteration method for the solution of the eigenvalue problem of
  linear differential and integral operators}.
\newblock \emph{J. Res. Natl. Bur. Stand. B}, 45:\penalty0 255--282, 1950.
\newblock \doi{10.6028/jres.045.026}.

\bibitem[LeCun et~al.(1992)LeCun, Simard, and Pearlmutter]{lecun1992}
Yann LeCun, Patrice Simard, and Barak Pearlmutter.
\newblock Automatic learning rate maximization by on-line estimation of the
  {H}essian\textquotesingle s eigenvectors.
\newblock In S.~Hanson, J.~Cowan, and C.~Giles, editors, \emph{Advances in
  Neural Information Processing Systems}, volume~5. Morgan-Kaufmann, 1992.
\newblock URL
  \url{https://proceedings.neurips.cc/paper_files/paper/1992/file/30bb3825e8f631cc6075c0f87bb4978c-Paper.pdf}.

\bibitem[Lewkowycz et~al.(2020)Lewkowycz, Bahri, Dyer, Sohl-Dickstein, and
  Gur-Ari]{lewkowycz2020large}
Aitor Lewkowycz, Yasaman Bahri, Ethan Dyer, Jascha Sohl-Dickstein, and Guy
  Gur-Ari.
\newblock The large learning rate phase of deep learning: the catapult
  mechanism, 2020.

\bibitem[Li et~al.(2018)Li, Xu, Taylor, Studer, and
  Goldstein]{li2018visualizing}
Hao Li, Zheng Xu, Gavin Taylor, Christoph Studer, and Tom Goldstein.
\newblock Visualizing the loss landscape of neural nets, 2018.

\bibitem[MacKay(1992)]{djcmThesis}
David~J.C. MacKay.
\newblock Bayesian methods for adaptive models, 1992.

\bibitem[Maddox et~al.(2020)Maddox, Benton, and Wilson]{maddox2020rethinking}
Wesley~J. Maddox, Gregory Benton, and Andrew~Gordon Wilson.
\newblock Rethinking parameter counting in deep models: Effective
  dimensionality revisited, 2020.

\bibitem[Nesterov(2014)]{Nesterov2013}
Yurii Nesterov.
\newblock \emph{Introductory Lectures on Convex Optimization: A Basic Course}.
\newblock Springer Publishing Company, Incorporated, 1 edition, 2014.
\newblock ISBN 1461346916.

\bibitem[Papyan(2019{\natexlab{a}})]{papyan2019measurements}
Vardan Papyan.
\newblock Measurements of three-level hierarchical structure in the outliers in
  the spectrum of deepnet {H}essians, 2019{\natexlab{a}}.

\bibitem[Papyan(2019{\natexlab{b}})]{papyan2019spectrum}
Vardan Papyan.
\newblock The full spectrum of deepnet {H}essians at scale: Dynamics with
  {S}{G}{D} training and sample size, 2019{\natexlab{b}}.

\bibitem[Papyan(2020)]{papyan2020traces}
Vardan Papyan.
\newblock Traces of class/cross-class structure pervade deep learning spectra.
\newblock \emph{Journal of Machine Learning Research}, 21\penalty0
  (252):\penalty0 1--64, 2020.
\newblock URL \url{http://jmlr.org/papers/v21/20-933.html}.

\bibitem[Pearlmutter(1994)]{pearlmutter}
Barak~A. Pearlmutter.
\newblock {Fast Exact Multiplication by the {H}essian}.
\newblock \emph{Neural Computation}, 6\penalty0 (1):\penalty0 147--160, 01
  1994.
\newblock ISSN 0899-7667.
\newblock \doi{10.1162/neco.1994.6.1.147}.
\newblock URL \url{https://doi.org/10.1162/neco.1994.6.1.147}.

\bibitem[Sagun et~al.(2017)Sagun, Bottou, and LeCun]{sagun2017eigenvalues}
Levent Sagun, Leon Bottou, and Yann LeCun.
\newblock Eigenvalues of the {H}essian in deep learning: Singularity and
  beyond, 2017.

\bibitem[Vaswani et~al.(2017)Vaswani, Shazeer, Parmar, Uszkoreit, Jones, Gomez,
  Kaiser, and Polosukhin]{vaswani2017attention}
Ashish Vaswani, Noam Shazeer, Niki Parmar, Jakob Uszkoreit, Llion Jones,
  Aidan~N. Gomez, Lukasz Kaiser, and Illia Polosukhin.
\newblock Attention is all you need, 2017.

\bibitem[Wang and Roberts(2023)]{sane}
L.~Wang and S.J. Roberts.
\newblock {S}{A}{N}{E}: the phases of gradient descent through sharpness aware
  number of effective pamareters.
\newblock 2023.

\bibitem[Xiao et~al.(2017)Xiao, Rasul, and Vollgraf]{xiao2017fashionmnist}
Han Xiao, Kashif Rasul, and Roland Vollgraf.
\newblock Fashion-{M}{N}{I}{S}{T}: a novel image dataset for benchmarking
  machine learning algorithms, 2017.

\bibitem[Ye and Lim(2016)]{ye2016schubert}
Ke~Ye and Lek-Heng Lim.
\newblock Schubert varieties and distances between subspaces of different
  dimensions, 2016.

\end{thebibliography}

\newpage

\begin{appendix}

\section{Experimental details}
In this section, we detail the technical details used in the experiments in the main sections.

\textbf{Lanczos iteration.} We can get the HVP using \citet{pearlmutter}'s trick, and we use the \citet{Lanczos:1950zz} algorithm for our Hessian $\mathcal{H}$ computations. Let the number of Lanczos iterations be $n_\mathrm{L}$, the algorithm returns a tridiagonal matrix $T \in \mathcal{R}^{n_\mathrm{L}} \times \mathcal{R}^{n_\mathrm{L}}$. We use $n_\mathrm{L}=100$, and the eigenvalues of the smaller $100 \times 100$ tridiagonal matrix can be readily computed using existing numerical libraries (e.g. \textit{numpy}). The eigenvectors of $\mathcal{H}$ can be computed easily, $V = V_T^T V_\mathrm{L}$, where $V_T$ are the eigenvectors of the tridiagonal matrix $T$ and $V_L$ the Lanczos vectors as secondary outputs from the algorithm. We perform re-orthogonalisation on the matrix of Lanczos vectors after every iteration. Our implementation of jax-powered \cite{jax2018github} Lanczos references a baseline implementation from \textit{https://github.com/google/spectral-density}. 

\textbf{FMNIST.} We train 5-layer MLPs with 32 hidden units in each layer for classification on the FMNIST dataset with cross-entropy loss. Our neural layers use ReLU activation, introduced by \citet{relu}. 
% In Fig. \ref{fig:fmnist-archs}, we train with MSE loss using one-hot encoding for classification. 
As pointed out by \citet{granziol2020lr}, the batch size $b$ of data can influence the sharpness of the landscape up until a regime of large $b$ where the eigenvalue from $\mathcal{H}_\mathrm{emp}$ dominates the scaling term. Following these intuitions, we compute optimal batch-size $b$ for FMNIST, and found that beyond $b=1000$, the Hessian at initialisation did not reduce significantly in sharpness. This implies that $\mathcal{H}_\mathrm{batch}$ is no longer dominated by the scaling term and so is a $\lambda_{i, b} \approx \lambda_{i, \mathrm{emp}}$. As a result, $b=1000$ represents our full training dataset on FMNIST. To ensure classes are well-represented in the training dataset, we construct the dataset with the first $4$ classes of FMNIST, so that each class will be represented by $\sim 250$ instances in the training dataset. The evaluation set is the same size, $b_\mathrm{eval}=1000$. For \grassd, we use $m=4$ equal to the number of classes in the classification problem. 

\textbf{CIFAR-10.} We train modified versions of AlexNet, introduced by \citet{alexnet}, on CIFAR-10 with cross-entropy loss. The network architecture uses $2$ sets of convolution \& max-pool blocks. Convolution layers are structured as ($64$ features, $(5,5)$ kernel, $(2, 2)$ strides) and max-pool as ($(3,3)$ kernel). This structure is followed by $2$ dense (fully-connected) layers with $382$ and $196$ hidden units respectively, and finally an output layer, following the modifications of \citet{keskar2017}. Our model uses ReLU activation \cite{relu}. Similar to FMNIST, we computed an optimal reduced batch size from the full training set, in this case, $b=5000$. All $10$ classes are used in this task, so each class is represented by $\sim 500$ instances in the training dataset. The evaluation set is smaller, $b_\mathrm{eval}=1000$. All models were trained to $1000$ epochs.

\newpage

\section{Phases of learning on FMNIST, higher $\eta$}
\begin{figure}[t] 
\centering
 \begin{subfigure}[b]{0.9\textwidth}
     \centering
     \includegraphics[width=0.9\textwidth]{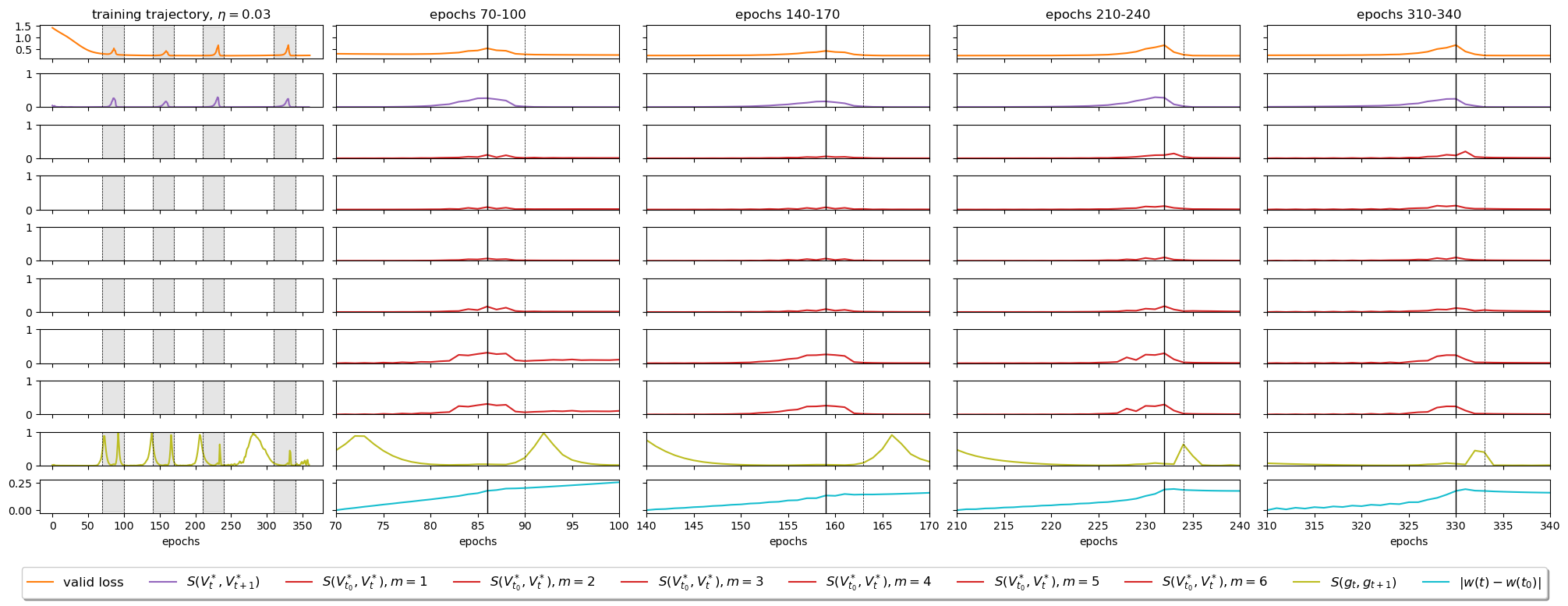}
     \caption{Loss, $S(V^*_t, V^*_{t+1}$ for different $m$ on FMNIST, $\eta=0.03$}
 \end{subfigure}
 \hfill
 \begin{subfigure}[b]{0.9\textwidth}
     \centering
     \includegraphics[width=0.9\textwidth]{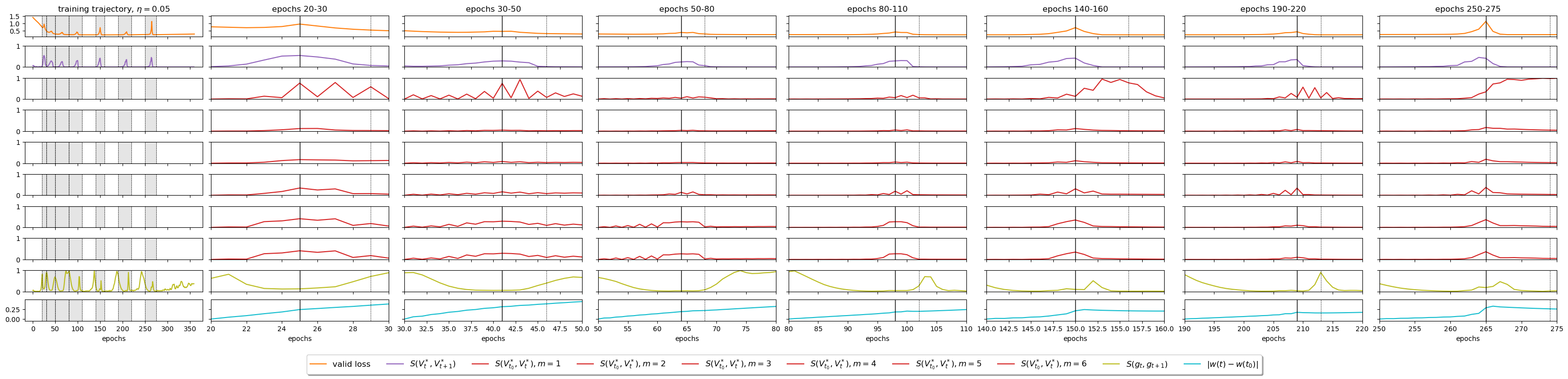}
     \caption{Loss, $S(V^*_t, V^*_{t+1}$ for different $m$ on FMNIST, $\eta=0.05$}
 \end{subfigure}
 \begin{subfigure}[b]{0.9\textwidth}
     \centering
     \includegraphics[width=0.9\textwidth]{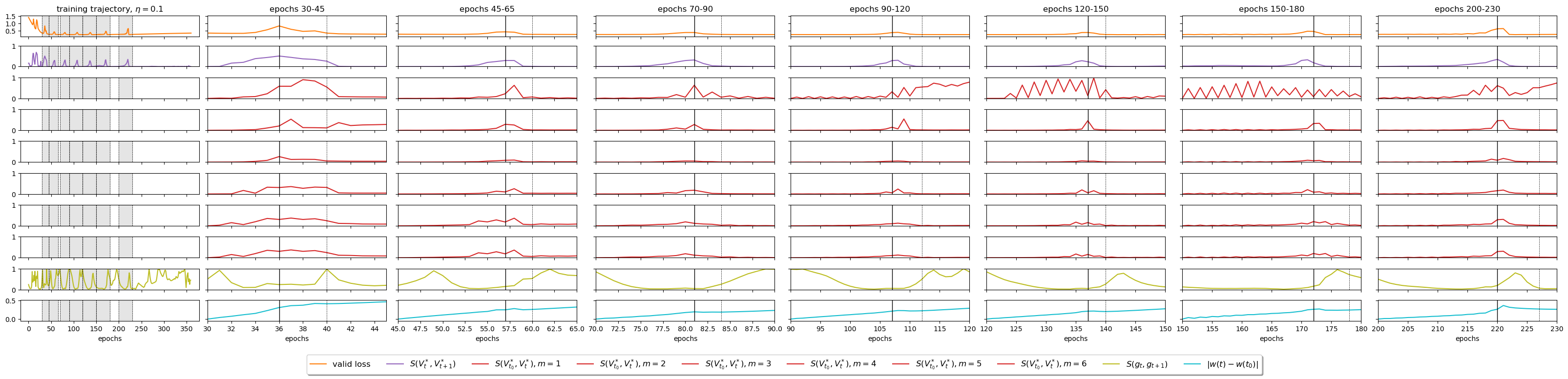}
     \caption{Loss, $S(V^*_t, V^*_{t+1}$ for different $m$ on FMNIST, $\eta=0.10$}
 \end{subfigure}

  \begin{subfigure}[b]{0.9\textwidth}
     \centering
     \includegraphics[width=0.9\textwidth]{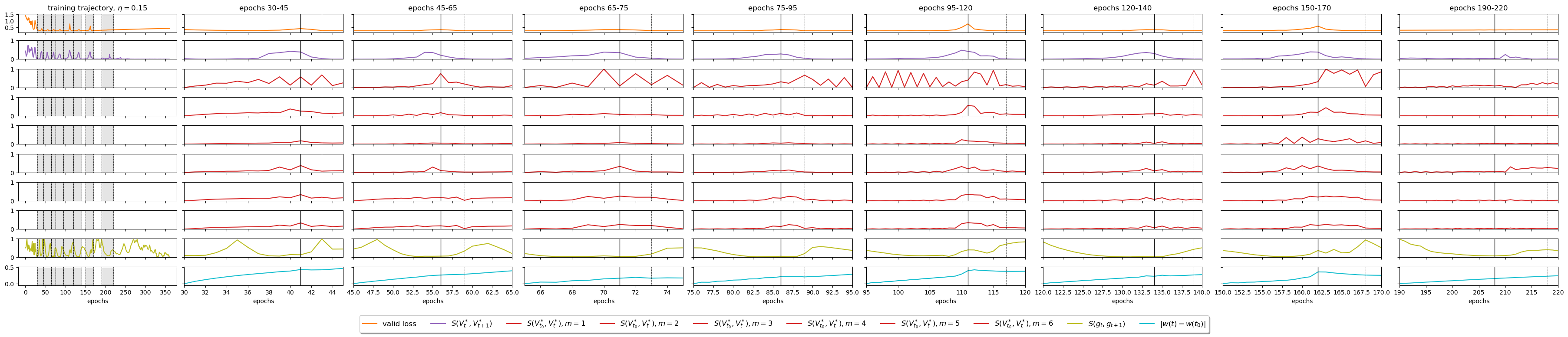}
     \caption{Loss, $S(V^*_t, V^*_{t+1}$ for different $m$ on FMNIST, $\eta=0.15$}
 \end{subfigure}
 
 \caption{\textbf{}
 } \label{fig:diff-m}

\end{figure}

Section \ref{section:main} presented our claims with detailed computations on FMNIST. In this section, we show the phases of learning at higher learning rates in Fig. \ref{fig:fmnist-more}, and we look at different choices of $m$ to compute $S(V^*_p, V^*_q)$ in Fig \ref{fig:diff-m}. The reordering effect is clearer for higher $\eta$, where high misalignments are observed for lower $m$.

\begin{figure}[t] 
\centering
 \begin{subfigure}[b]{0.9\textwidth}
     \centering
     \includegraphics[width=0.9\textwidth]{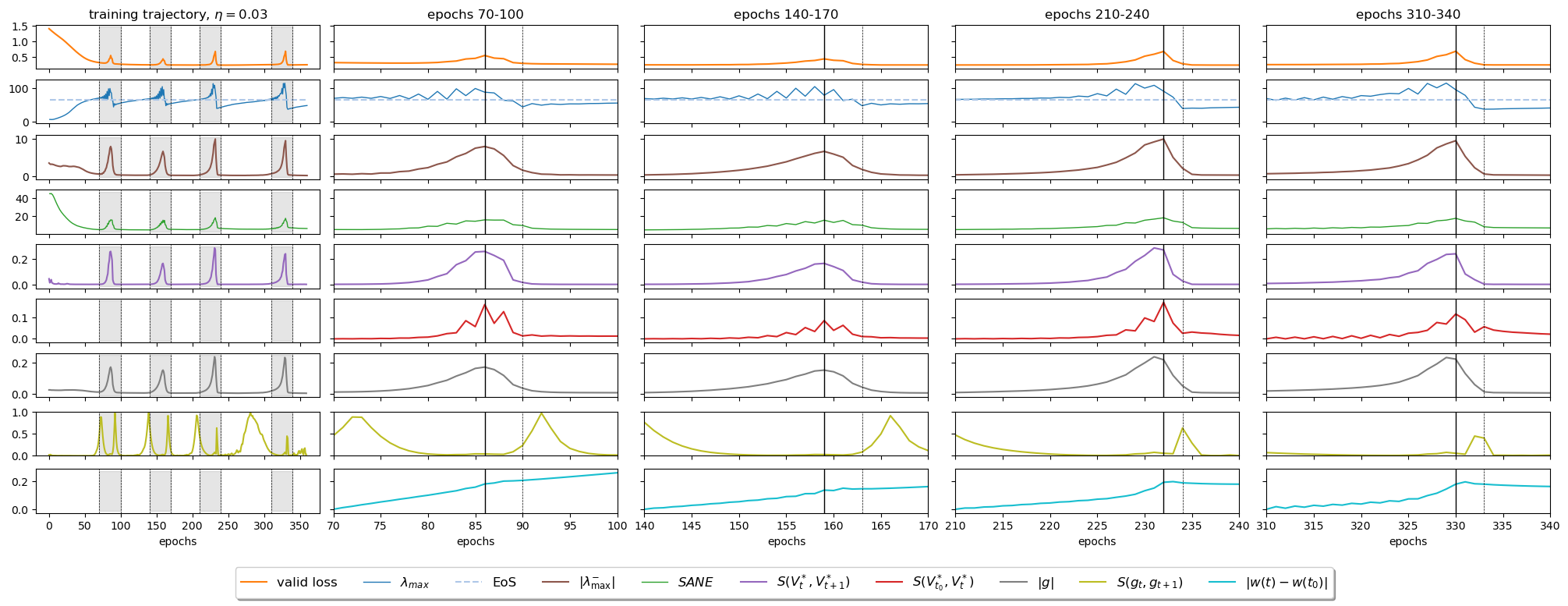}
     \caption{Training trajectory on CIFAR-10, $\eta=0.06$}
 \end{subfigure}
 \hfill
 \begin{subfigure}[b]{0.9\textwidth}
     \centering
     \includegraphics[width=0.9\textwidth]{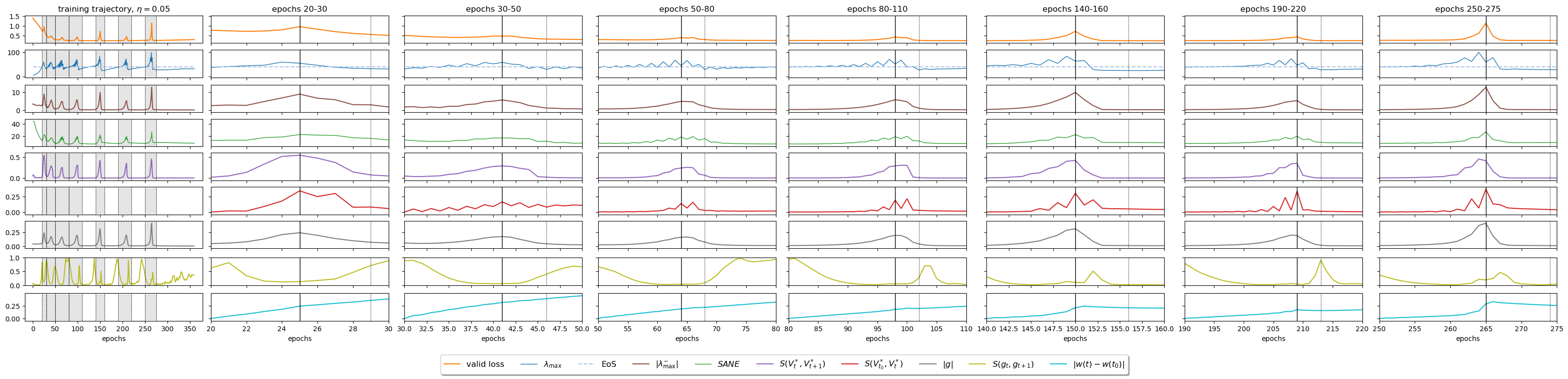}
     \caption{Training trajectory on CIFAR-10, $\eta=0.06$}
 \end{subfigure}
 \begin{subfigure}[b]{0.9\textwidth}
     \centering
     \includegraphics[width=0.9\textwidth]{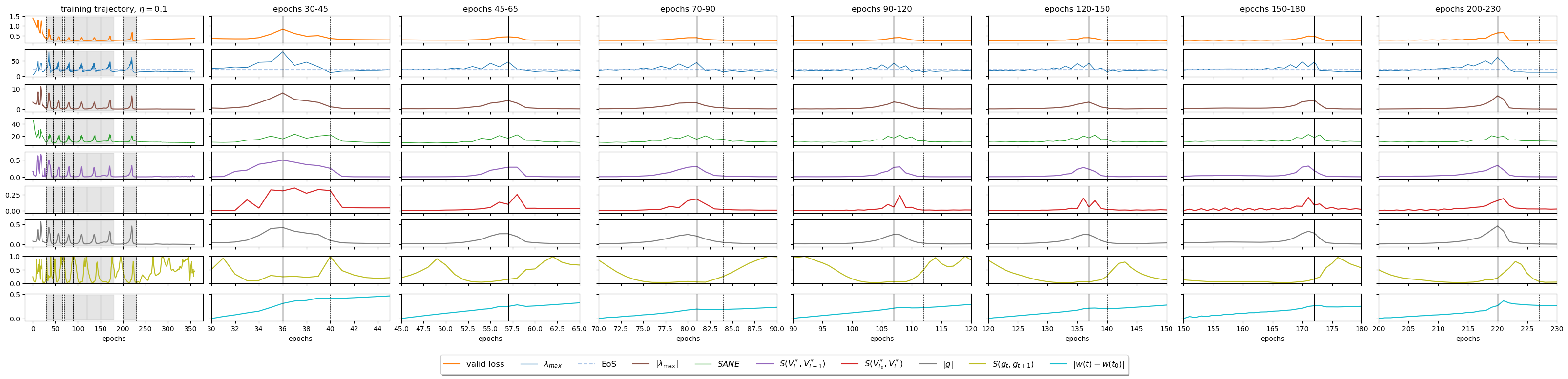}
     \caption{Training trajectory on CIFAR-10, $\eta=0.06$}
 \end{subfigure}

  \begin{subfigure}[b]{0.9\textwidth}
     \centering
     \includegraphics[width=0.9\textwidth]{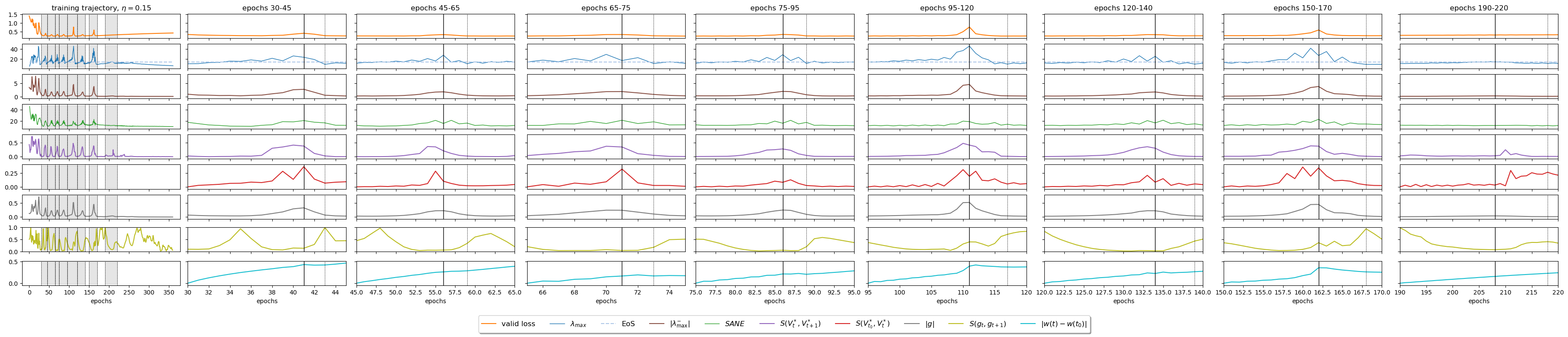}
     \caption{Training trajectory on CIFAR-10, $\eta=0.06$}
 \end{subfigure}
 
 \caption{\textbf{}
 } \label{fig:cifar}

\end{figure}

\newpage

\newpage
\section{Phases of learning on CIFAR-10}\label{sm:cifar}
Section \ref{section:main} presented our claims with detailed computations on FMNIST. In this section, we show the phases of learning on CIFAR-10 \cite{krizhevsky2009learning}, a more challenging dataset for image classification. To scale our computation to deep neural nets, we present an approximation to the full Hessian that exploits the natural ordering of neural weights. We justify this approximation in detail in \app\ section \ref{sm:synthetic}

\begin{figure}[h] 
\centering
 \begin{subfigure}[b]{1.0\textwidth}
     \centering
     \includegraphics[width=0.9\textwidth]{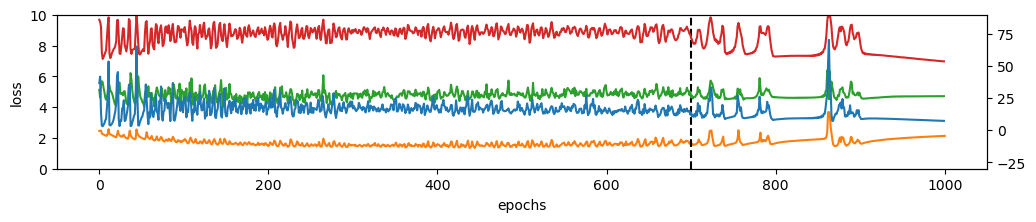}
     \caption{Training trajectory on CIFAR-10, $\eta=0.06$}
        \label{fig:cifartop}
 \end{subfigure}
 \hfill
 \begin{subfigure}[b]{1.0\textwidth}
     \centering
     \includegraphics[width=0.9\textwidth]{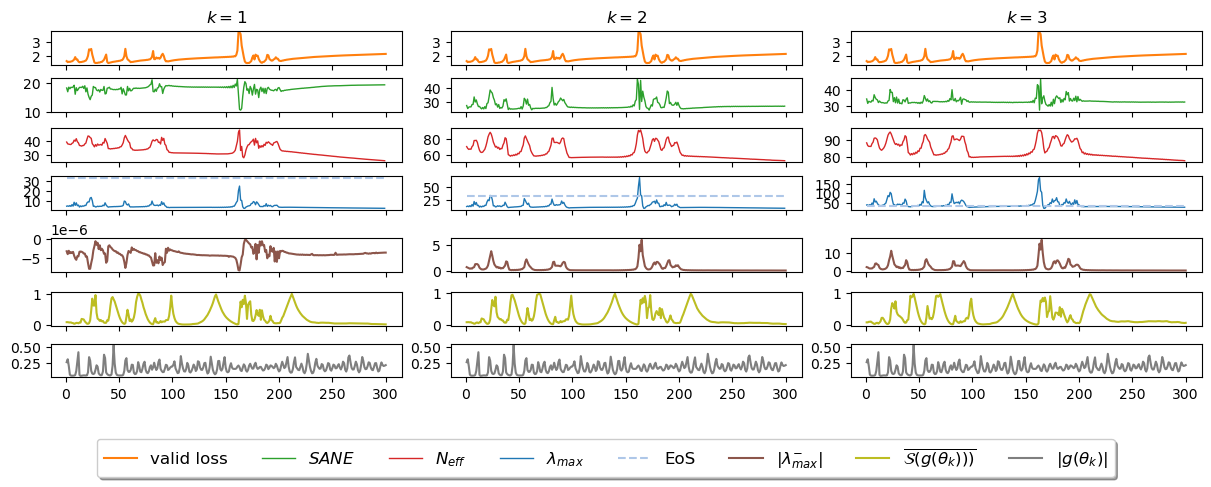}
     \caption{Validation loss and measures; zoomed in to epochs 700-1000}
        \label{fig:cifarbot}
 \end{subfigure}
 
 \caption{\textbf{Training on CIFAR-10 displays phases of instability toward the end}. \textbf{Top:} training trajectory on CIFAR-10 for 1000 epochs. Validation loss (orange and below) plotted on the left axis, while \neff\, \sane\, \lammax\, (in descending order) plotting on the right axis. Axes have been rescaled to promote visibility. \textbf{Bottom:} Plots of validation loss vs measures computed from \redhess\ from epoch $700$, at varying depths $k=1, 2, 3$. We note that the phases of gradient descent are visible from epoch $700$. Additionally, the gradient displays remarkable similarity across the depths of our approximation.
 } \label{fig:cifar}

\end{figure}

\noindent \textbf{Approximation to the full Hessian. }
We parameterise a prediction function $\hat{y_i} = f_\theta(x_i)$ with a deep neural net with weights $\theta$. Let the network have layers and outputs ($\phi_k$, $T_k$), $k = 1, 2, ... n_\phi$, ordered from the output layer such that $\phi_{k+1}(T_{k+1}, \theta_{k+1}) = T_k$ and $T_{n_\phi}=\textbf{x}, T_0=f_\theta(\textbf{x})$, then $f_\theta(x_i)=\phi_1(\phi_{2}(...(\phi_{n_\phi}(x_i))))$. We consider the reduced objective $\mathcal{L}^\mathrm{r}_{\theta(k)}=\mathbb{E}_{\bm{T}_k} L_{\theta(k)}(\bm{T}_k)$, where $\theta(k)=[\theta_1, \theta_2, ..., \theta_k]$ to compute the reduced Hessian $\mathcal{H}^\mathrm{r}_k=\grad^2_{\theta(k)} \mathcal{L^\mathrm{r}}_{\theta(k)}$. In other words, we approximate the Hessian of the full deep neural network with a Hessian computed on the first $k$ layers closest to the output layer. Empirically, this approximation changes the absolute scale of the resulting eigen-spectrum, but the relative scales of measures along the training trajectory are accurate. 

\noindent \textbf{Experiments on CIFAR-10. } Fig. \ref{fig:cifar} shows a specific training trajectory, where we initially observe a large amount of noise. After epoch 700, the trajectory displays the structure of phase transitions, and we visualise the details in Fig. \ref{fig:cifarbot} across depths $k=1,2,3$, where $k=1$ is the output layer only. For each depth, the percentage of model parameters included in the computation was $0.2\%, 9.7\%, 86.5\%$, showing that a more accurate approximation of the full Hessian as $k$ is increased. We note the significant difference in $k=1$ computations vs $k=2,3$, and we note that the output layer is unique in its lack of an activation function. Sadly, we were not able to compute \rotins\ given the limit of GPU memory. We note the clear phases as shown by loss and the measures \sane\, \neff\, and \lammax\, with the corresponding peaks and troughs as described in Section \ref{section:main}. However, the insight of Hessian rotations through \rotins\ and $|g(\theta_k)|$ is not clear, though we note the remarkable similarity of $|g(\theta_k)|$ and \gins\ across depths.

% \section{Related works}

\section{Additional synthetic experiments}\label{sm:synthetic}
\begin{figure}[h] 
% \centering
 \begin{subfigure}[b]{0.24\textwidth}
     \centering
     \includegraphics[width=\textwidth]{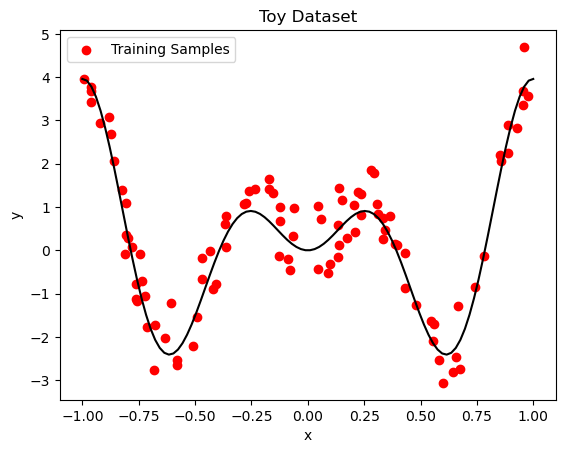}
     \caption{W-reg, train set}
 \end{subfigure}
 \begin{subfigure}[b]{0.24 \textwidth}
     \centering
     \includegraphics[width=\textwidth]{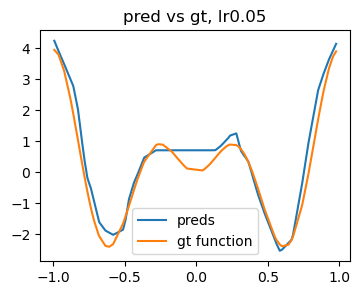}
     \caption{W-reg, sample preds}

 \end{subfigure}
 \begin{subfigure}[b]{0.24 \textwidth}
     \centering
     \includegraphics[width=\textwidth]{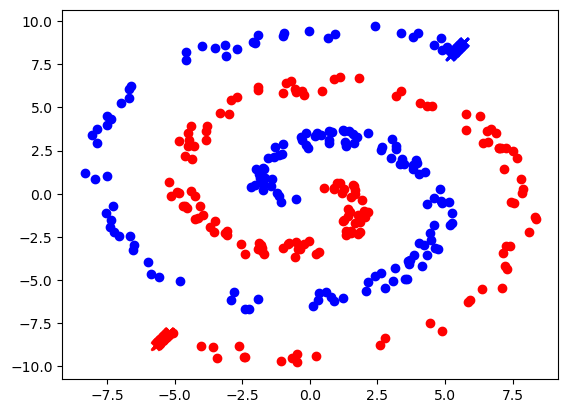}
     \caption{SRC, train set}
 \end{subfigure}
 \begin{subfigure}[b]{0.24 \textwidth}
     \centering
     \includegraphics[width=\textwidth]{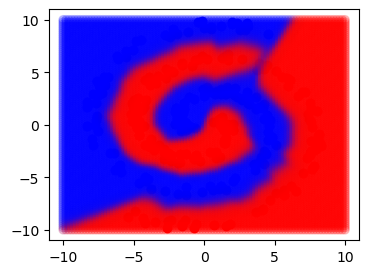}
     \caption{SRC, sample preds}

 \end{subfigure}
 
 \caption{\textbf{Synthetic datasets, \textit{W-reg} \& \textit{SRC}.} Training dataset and sample predictions.} \label{fig:additional-preds}
 
\end{figure}

In this section, we establish the connection of eigenvectors to specific degrees-of-freedom that control performance and generalisation, and we visualise the behaviour of the $k$-layer approximation to the full Hessian introduced in Section \ref{sm:cifar}. Owing to the excessive amounts of compute required for larger datasets and models, we validate these observations on two small synthetic datasets - one for regression and one for classification. Synthetic datasets have the added benefit of allowing a lower dimensional input space to enable more intuitive visualisations of regression predictions and classification boundaries. The regression task fits the function $f(x) = 4x \sin(8x)$, which makes a \textit{W}-shape in the domain $[-1, 1]$, we call this dataset \textbf{W-reg}. For classification, we use a {S}wiss-{R}oll (\textbf{SRC}) dataset which represents a complex transformation from the two-dimensional input space to the feature space. These synthetic datasets are plotted in Fig. \ref{fig:additional-preds}. 

\begin{figure}[p] 
\centering
\makebox[\textwidth][c]{
\begin{subfigure}[b]{0.9\textwidth}
     \includegraphics[width=\textwidth]{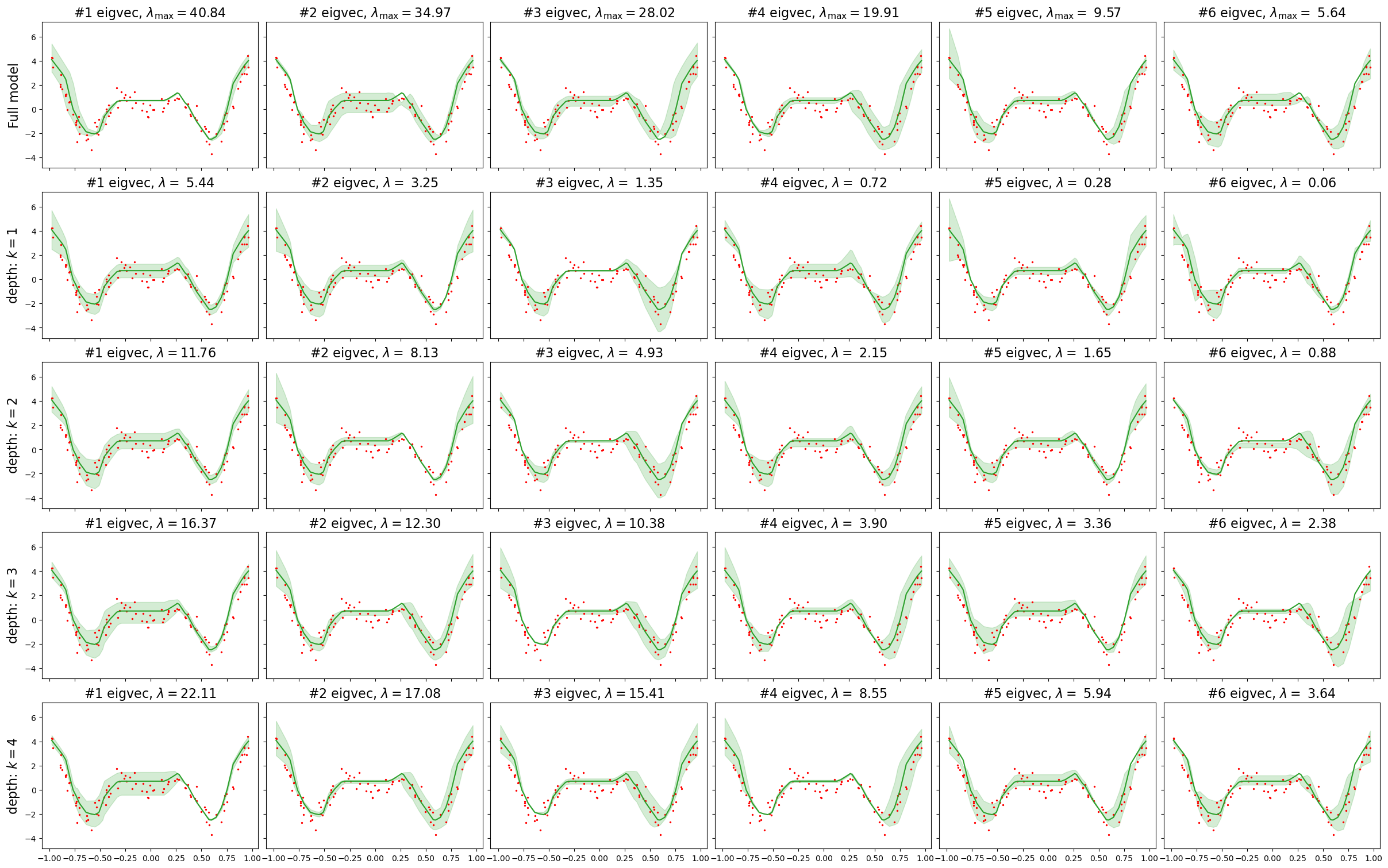}
\end{subfigure}}

 \makebox[\textwidth][c]{
\begin{subfigure}[b]{0.9\textwidth}
     \includegraphics[width=\textwidth]{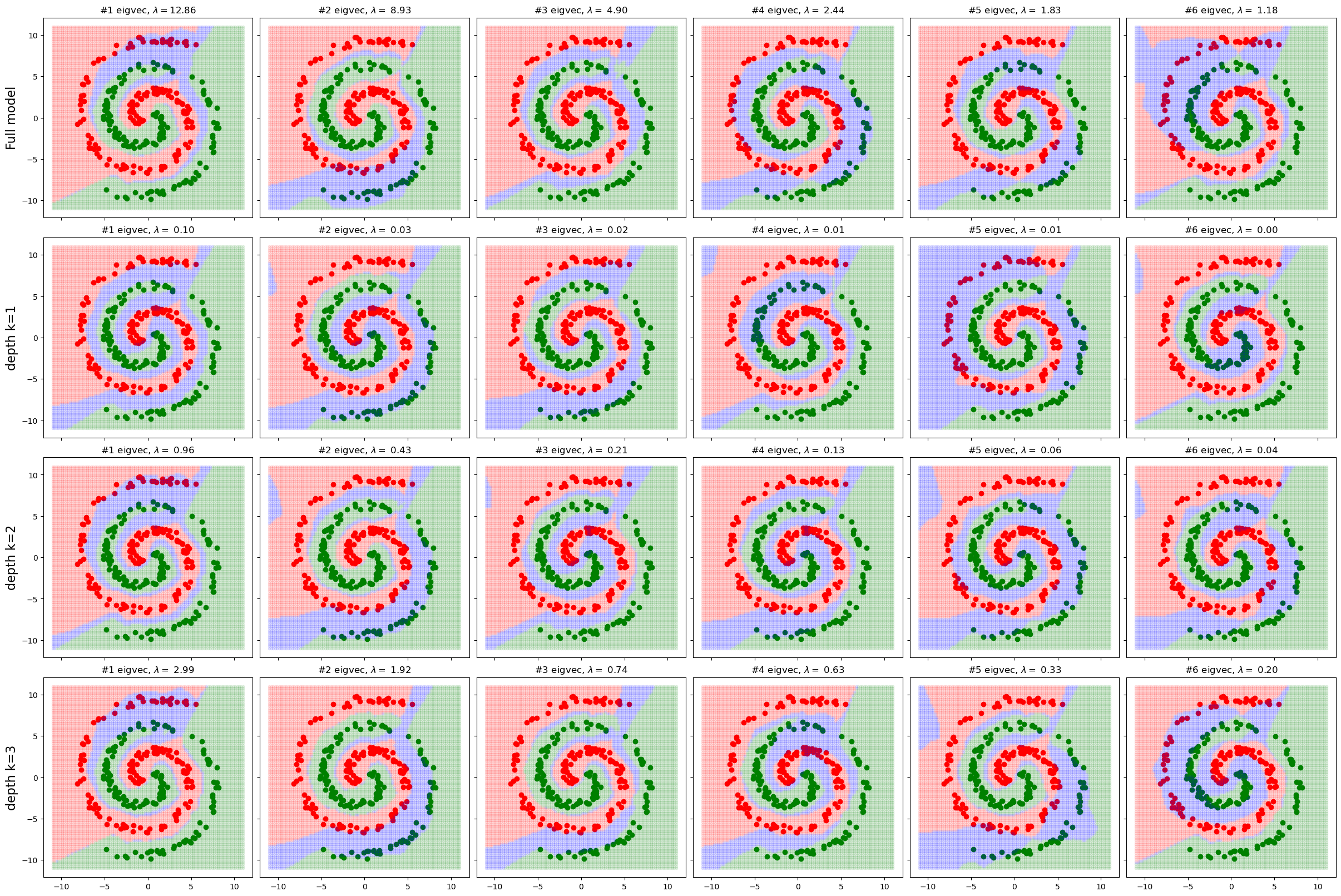}
\end{subfigure}
 }
 
\caption{\textbf{Perturbation studies of top $n=6$ Hessian eigenvectors. } \textbf{Top:} Epoch $2000$ of a $5$-layer MLP trained on \textbf{W-reg}, using $\eta=0.05$, and $k=1, 2, 3, 4$. \textbf{Bottom:} Epoch $2000$ of a $4$-layer MLP trained on \textbf{SRC}, using $\eta=0.05$, and $k=1, 2, 3$. We note that perturbations on eigen-directions of Hessians correspond to uncertainties in local regions of the solution. These eigen-features change minimally when we move the domain of the Hessian from the output layer deeper into the neural net. } \label{fig:regsrc-k}
 
\end{figure}

\subsection{"Effective" parameters control important degrees-of-freedom}
\textbf{Perturbations.} We train MLPs for both W-reg and SRC using MSE and cross-entropy loss at large $\eta$s. Taking models at the final epoch, we visualise the specific degrees-of-freedom (\dofs) corresponding to the top $n=6$ eigenvectors of the Hessian. These \dofs\ are evaluated through perturbation theory, allowing us to perturb the model weights along the eigen-directions and measure the effect on output space. The weights are perturbed as $\theta_{p,i} = \theta \pm p_i v_i$, where $p$ is a scaling factor. Perturbations on eigen-directions of the full-model Hessian, as well as with $k$-layer approximations of varying depths, are shown in Fig. \ref{fig:regsrc-k}. In a full-batch and large $\eta$ regime, we expect the sharpness of final models to be determined by the unstable dynamics of gradient descent and by the \eos. As $\lambda$s get increasingly sharp, the landscape in the two-dimensional plane defined by $[v, \mathcal{L}]$ gets increasingly sharp, and so we use a square-root scaling factor (based on the quadratic assumption) $p_i=|\lambda_i|^{\frac{1}{2}} c_p$. The positive and negative perturbations form the boundaries of the error bars of the visualisations.

\noindent \textbf{W regression.} Focusing on the top subplot of the top plot of Fig. \ref{fig:regsrc-k}, we can qualitatively attribute the components of the regression solution to the specific eigenvectors. From the edges, $v_1$ controls the local \dof\ of the left edge of the desired output, and $v3$ is the right. Moving inwards, the second bend from both the left and right are controlled by $v_2$ and $v_4$ respectively. $v_2$ also controls the middle plateau, and $v_5$, $v_6$ offer more precise tuning for the sharp turns at the bottom of the \textit{W} shape. We find it surprising that the sharpest eigen-directions appear to control salient \dofs that correspond to performance in the \textit{local} region, and that these sharpest eigenvectors form a 'sum of local parts' to generate the whole solution. We note that the ordering of $v_1$, $v_2$, and $v_3$ is coincidentally similar to the frequency of datapoints in the training set within the respective regions of the input domain. Since the empirical Hessian is driven by the loss from training samples, it is possible that the relative sharpness of $v$s are determined by the frequency of the corresponding feature in the training set. 

\noindent \textbf{Swiss-{R}oll classification.} The perturbations plots for SRC are shown in the top subplot of the bottom plot of Fig. \ref{fig:regsrc-k}. The red and green show the 'unperturbed' decision boundaries, while the error bars on the classification boundary due to perturbation have a blue fill. Given the more landscape compared to W-reg, we note that the sharpest eigen-directions of the Hessian correspond to features that are local ($v_1$, $v_2$, $v_6$, and arguably $v_3$). $v_4$ and $v_5$ focus on the boundaries between the swirls - while they look similar in shape, the regions of uncertainty for each feature differ and are complementary.

\subsection{The $k$-layer approximation of the Hessian maintains relative scaling}
\begin{figure}[h] 
% \centering
 \begin{subfigure}[b]{\textwidth}
     \centering
     \includegraphics[width=0.8\textwidth]{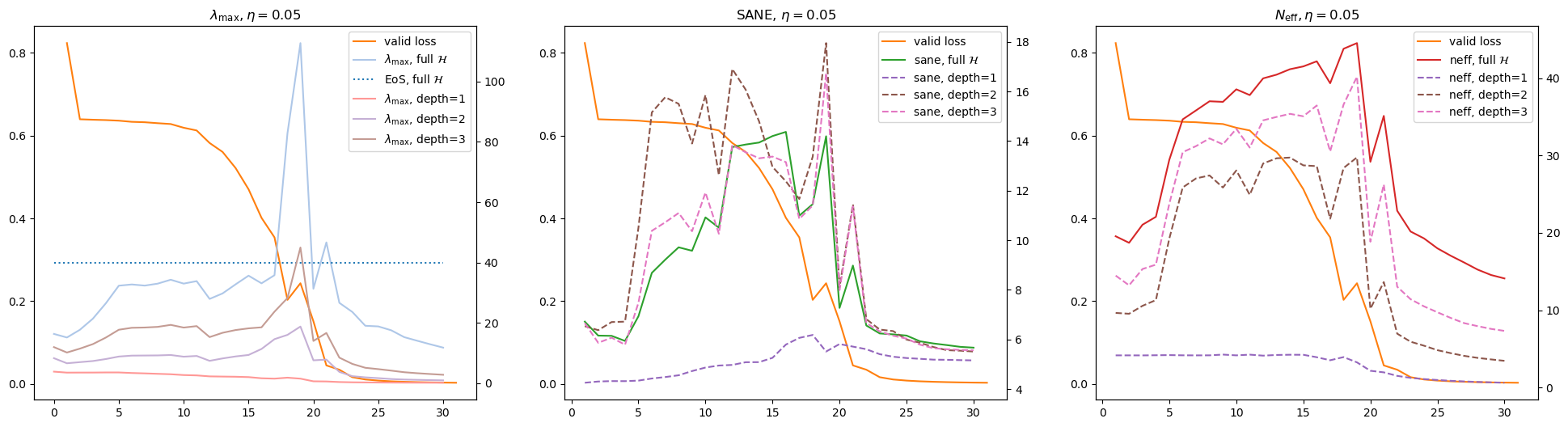}
     % \caption{some text}
 \end{subfigure}
 \begin{subfigure}[b]{\textwidth}
     \centering
     \includegraphics[width=0.8\textwidth]{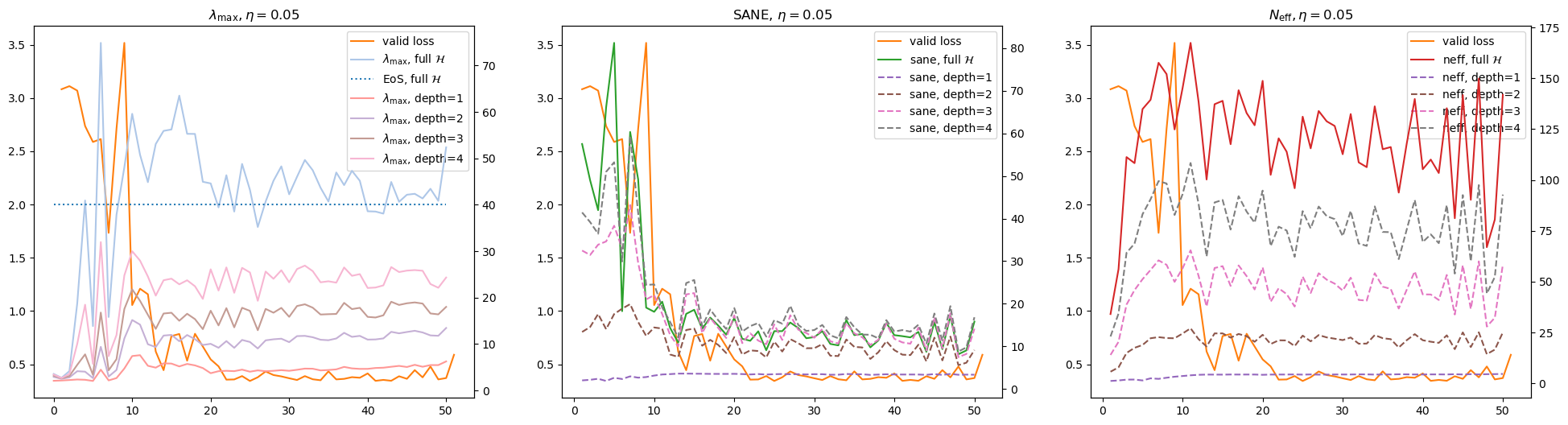}
     % \caption{some text}

 \end{subfigure}
 
 \caption{\textbf{\sane, \neff, and \lammax\ computed from the Hessian taken at different depths.} \\ \textbf{Top:} \textit{W-reg}. \textbf{Bottom:} \textit{SRC}. We see that the metrics are increasingly similar in relative scale, and the absolute scales are increasing close to the full model as $k$ is increased. The $k=1$ approximation, which uses only the output-layer, produces flat and uninformative metrics. } \label{fig:metrics-k}
 
\end{figure}

In section \ref{sm:cifar}, we introduced the $k$-layer approximation to the Hessian \redhess\ that exploits the natural ordering of weights. In Fig. \ref{fig:metrics-k} we compare metrics (\sane, \neff, \lammax) computed from \redhess\ to those from $\mathcal{H}$, the full Hessian, on synthetic datasets. In Fig. \ref{fig:regsrc-k}, we perturb the eigenvectors computed from \redhess\ at different depths. From the empirical evidence, we observe that the relative scales of the metrics computed from \redhess\ follow metrics from the full Hessian. This observation was utilised in Section \ref{sm:cifar} to provide a connection between metrics from \redhess\ to model performance since only the relative scales along the trajectory are critical. Secondly, we note that the absolute scale of metrics computed from \redhess\ approaches those from $\mathcal{H}$ as $k$ is increased, which agrees with intuitions. As with Section \ref{section:main}, more work is required for an interpretation of the absolute values of \sane\, \neff, and \lammax. Thirdly, we note that \redhess\ using only the output layer, i.e. $k=1$, computes metrics that are highly uninformative. The low \lammax\ indicates a flat landscape. Despite this, the eigenvectors from \redhess\ $k=1$ correspond to local \dofs that are very similar to that of the full model, and we conjecture that the eigenvectors of the output layer exert significant control on the specific \dofs for eigenvectors approximated with more layers or from the full model. 

\newpage

% \newpage
\section{Additional studies}

\subsection{Hessian shifts with cyclic learning rates}
\begin{figure}[h] 
% \makebox[\textwidth][c]{
\centering
 \begin{subfigure}[l]{0.2\textwidth}
     \centering
     \includegraphics[width=0.99\textwidth]{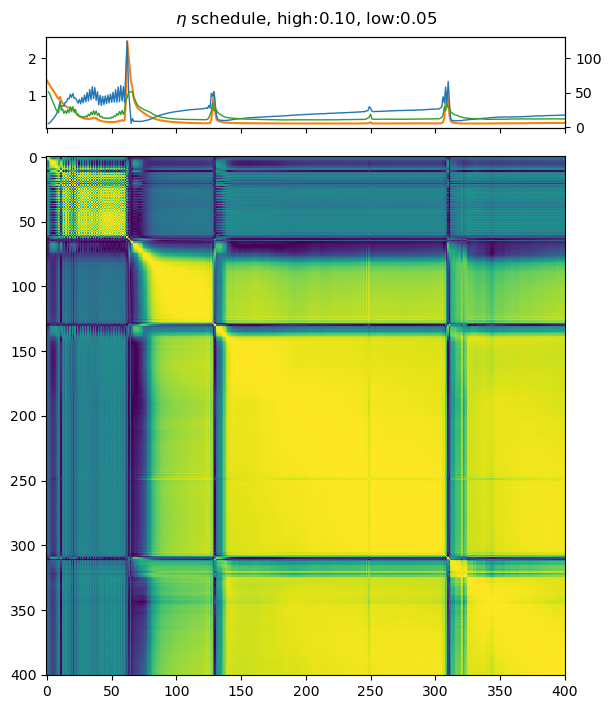}

 \end{subfigure}
 \begin{subfigure}[c]{0.2\textwidth}
     \centering
     \includegraphics[width=0.99\textwidth]{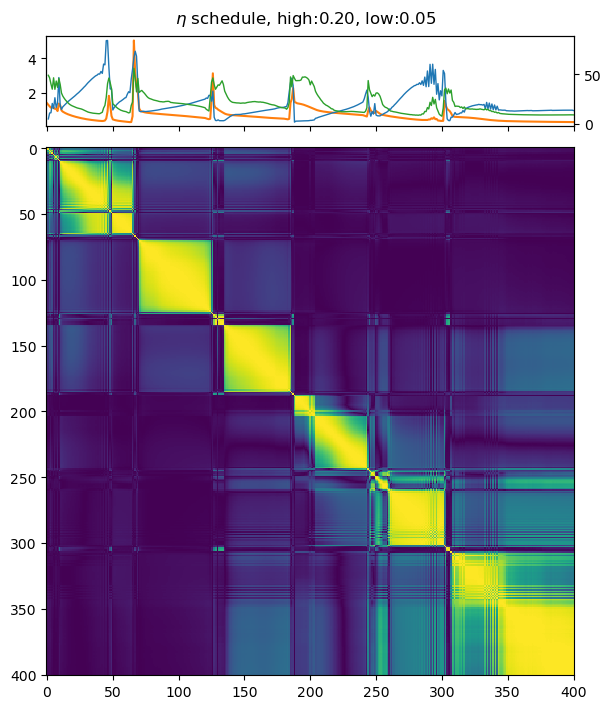}

 \end{subfigure}
 \begin{subfigure}[r]{0.2\textwidth}
     \centering
     \includegraphics[width=0.99\textwidth]{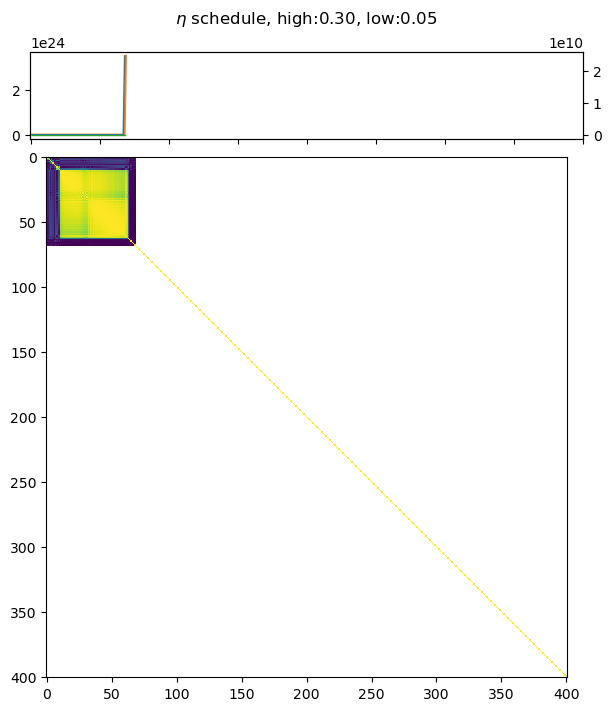}

 \end{subfigure}

\centering
 \begin{subfigure}[l]{0.2\textwidth}
     \centering
     \includegraphics[width=0.99\textwidth]{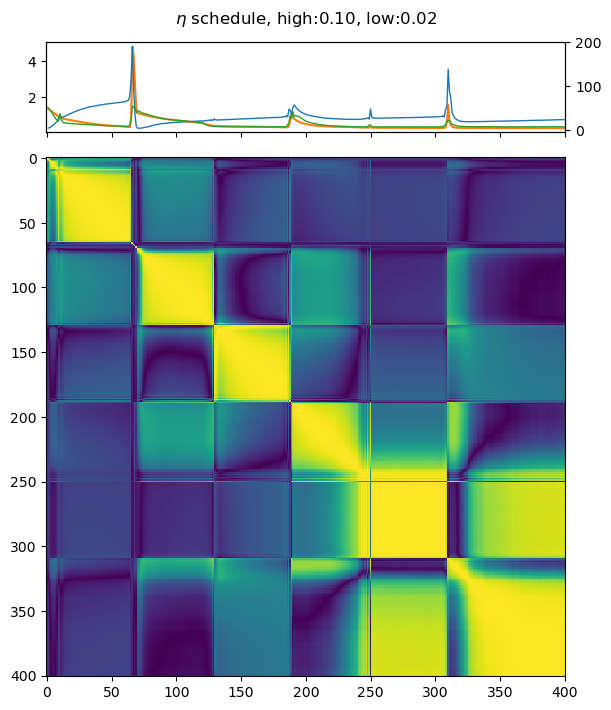}

 \end{subfigure}
 \begin{subfigure}[c]{0.2\textwidth}
     \centering
     \includegraphics[width=0.99\textwidth]{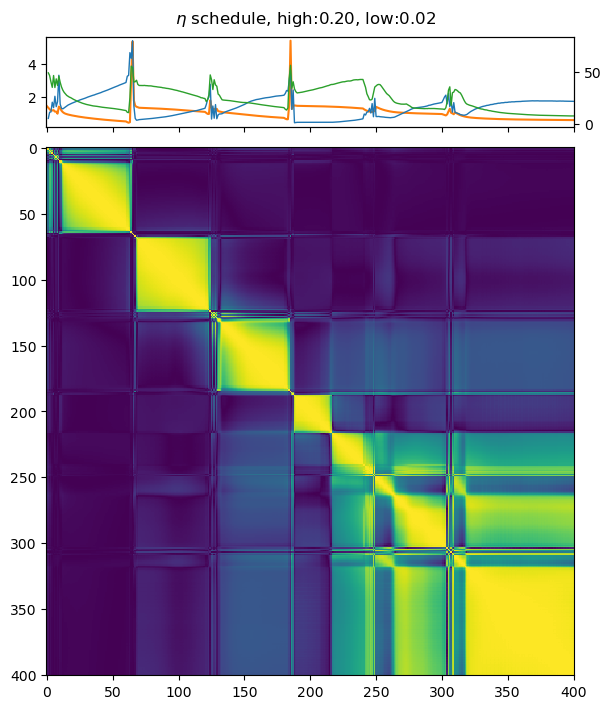}

 \end{subfigure}
 \begin{subfigure}[r]{0.2\textwidth}
     \centering
     \includegraphics[width=0.99\textwidth]{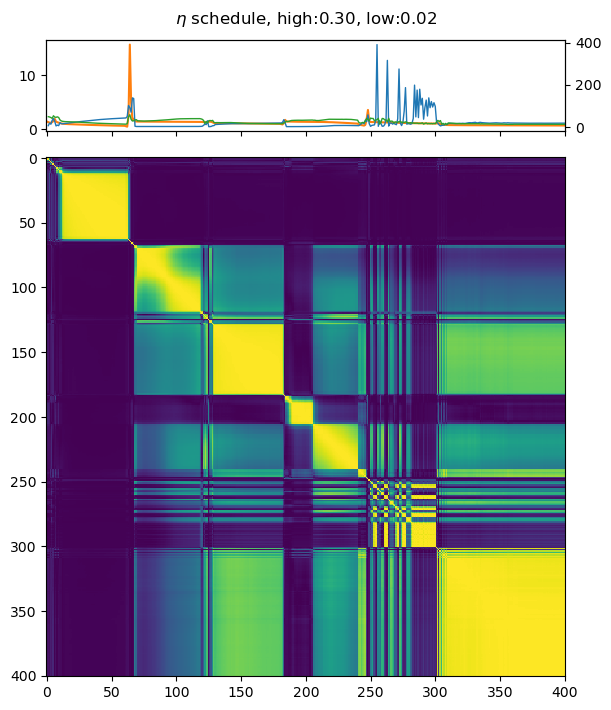}

 \end{subfigure}
 % }
  \hfill
 \caption{\textbf{Training trajectories and similarities of six cyclic $\eta$ schedules.} \textbf{Columns:} $\eta_+ \in (0.10, 0.20, 0.30)$. \textbf{Rows:} $\eta_- \in (0.05, 0.02)$. \textbf{Top of subplots:} learning trajectory (validation loss, \sane\, \neff\, \lammax\ plotted). \textbf{Bottom of subplots:} $S_c(v_{\mathrm{max},t_i}, v_{\mathrm{max},t_j})$ of cyclic $\eta$ schedules. The schedule of $\eta_\mathrm{high}=0.30$ and $\eta_\mathrm{low}=0.05$ has diverged during training. \textbf{Cyclic $\eta$ schedules can balance the exploration and exploitation trade-off of \textit{loss basins} depending on selected $\eta$s.} } \label{fig:cyclic}
 
\end{figure}

We study the exploration of \textit{loss landscapes} under six cyclic $\eta$ schedules through the similarity of $v_\mathrm{max}$. We take $\eta_+ \in (0.10,0.20,0.30)$, situated well within the \textit{unstable} regime, as the upper limits of our schedule; $\eta_- \in (0.02, 0.05)$ are used as the lower limits. $\eta=0.02$ is in the \gf\ earning regime while $\eta=0.05$ is \textit{unstable}. These cyclic $\eta$ schemes use $\eta_+$ for $10$ epochs, before switching to $\eta_-$ for $50$ epochs and repeating. The final $40$ epochs use $\eta_-$. The results are visualised in Fig. \ref{fig:cyclic}, and we see difference choices of $\eta_+$ and $\eta_-$ can lead to different movements of the Hessian. We observe that large $\eta_+$s encourage Hessian shifts. Surprisingly, it is not the case that using low $\eta_+$ and $\eta_-$ will lead to stagnation in similar \textit{loss landscapes}. The ratio $\frac{\eta_+}{\eta_-}$ appears to play an important role.

\end{appendix}

\end{document}